\pdfoutput=1
\documentclass[11pt]{article}

\usepackage{acl}
\usepackage{times}
\usepackage{latexsym}
\usepackage[T1]{fontenc}
\usepackage[utf8]{inputenc}
\usepackage{microtype}
\usepackage{booktabs}
\usepackage{multirow}
\usepackage{fontawesome}
\usepackage{bm} % bold kappa
\usepackage{calc}
\usepackage{colortbl}
\usepackage{blindtext}
\definecolor{level1color}{HTML}{E6F3FF}
\definecolor{level2color}{HTML}{B3D9FF}
\definecolor{level3color}{HTML}{7FBFFF}
\definecolor{level4acolor}{HTML}{7FC31C}
\definecolor{level4bcolor}{HTML}{800080}

\usepackage[pdftex]{graphicx}
\usepackage{tabularx}
\usepackage{subcaption}
\DeclareGraphicsExtensions{.pdf,.ai,.jpg,.png}
\setkeys{Gin}{pagebox=artbox}% Use artbox instead mediabox default. Possible values: mediabox, cropbox, bleedbox, trimbox, artbox. (shell: pdfinfo -box <pdf-file>)
\graphicspath{{./figures/}}

\usepackage{tikz}
\usetikzlibrary{decorations.text}

\usepackage{amssymb}
\usepackage{amsmath}

\RequirePackage{type1cm}
\RequirePackage{color}
\RequirePackage{soul}
\setstcolor{blue}
\definecolor{violet}{rgb}{0.5,0.0,0.5}
\newsavebox\bscombox
\newcommand{\bscom}[3][]{%
	\sbox{\bscombox}{\fontsize{8}{9}\selectfont#1#2#3}
	\noindent
	\st{#2}{\selectfont
		\color{blue}#3\ifx\\#1\\\else{\fontsize{8}{9}\selectfont\color{violet}[#1]}\fi
	}
}

\begin{document}

\setlength\titlebox{9cm}
\title{A Dataset for Identifying the Human Values behind Arguments}
\title{The ValueEval'23 Dataset for Identifying Human Values behind Arguments}
\title{The Touch{\'e}23-ValueEval Dataset for\\Identifying Human Values behind Arguments}

\author{%
  Nailia Mirzakhmedova\thanks{\hspace{1em}Contact: {nailia.mirzakhmedova@uni-weimar.de}} \\
  Bauhaus-Universit\"at Weimar \\
  % \texttt{nailia.mirzakhmedova@uni-weimar.de} \\%
  % ORCID: 0000-0002-8143-1405
\And
  Johannes Kiesel \\
  Bauhaus-Universit\"at Weimar \\
  % \texttt{johannes.kiesel@uni-weimar.de} \\%
  % ORCID: 0000-0002-1617-6508
\And
  Milad Alshomary \\
  Leibniz University Hannover \\
  % \texttt{m.alshomary@ai.uni-hannover.de} \\ %
  % ORCID: 0000-0001-6142-9124
\AND
  Maximilian Heinrich \\
  Bauhaus-Universit\"at Weimar \\
  % \texttt{maximilian.heinrich@uni-weimar.de} \\%
  % ORCID: 0000-0001-5450-8203
\And
  Nicolas Handke \\
  Universit\"at Leipzig \\
  % \texttt{n.handke@studserv.uni-leipzig.de} \\%
  % ORCID: 0000-0003-1349-4671
\And
  Xiaoni Cai \\
  Technische Universit\"at M\"unchen \\
  % \texttt{caix@in.tum.de} \\%
  % ORCID: 0000-0003-2114-8162
\AND
  Valentin Barriere \\
  CENIA \\
  % \texttt{valentin.barriere@cenia.cl} \\%
\And
  Doratossadat Dastgheib \\
  Shahid Beheshti University \\
  % \texttt{d_dastgheib@sbu.ac.ir}
\And
  Omid Ghahroodi\\
  Sharif University of Technology \\
  % \texttt{omid.ghahroodi98@sharif.edu}
  % ORCID: https://orcid.org/0000-0001-5577-882X
\AND
  Mohammad Ali Sadraei\\
  Sharif University of Technology\\
  % \texttt{m.sadraei@sharif.edu}
\And
  Ehsaneddin Asgari\\
  University of California Berkeley\\
  % \texttt{asgari@berkeley.edu}
\AND
  Lea Kawaletz \\
  Heinrich-Heine-Universit\"at\\D\"usseldorf \\
  % \texttt{lea.kawaletz@hhu.de} \\%
  % ORCID: 0000-0002-1473-2486
\And
  Henning Wachsmuth \\
  Leibniz University Hannover \\
  % \texttt{h.wachsmuth@ai.uni-hannover.de} \\ %
  % ORCID: 0000-0003-2792-621X
\And
  Benno Stein \\
  Bauhaus-Universit\"at Weimar \\
  % \texttt{benno.stein@uni-weimar.de} \\%
  % ORCID: 0000-0001-9033-2217
}

\maketitle

\begin{abstract}
We present the Touch{\'e}23-ValueEval Dataset for Identifying Human Values behind Arguments. To investigate approaches for the automated detection of human values behind arguments, we collected 9324 arguments from 6~diverse sources, covering religious texts, political discussions, free-text arguments, newspaper editorials, and online democracy platforms. Each argument was annotated by 3 crowdworkers for 54~values. The Touch{\'e}23-ValueEval dataset extends the Webis-ArgValues-22. In comparison to the previous dataset, the effectiveness of a 1-Baseline decreases, but that of an out-of-the-box BERT model increases. Therefore, though the classification difficulty increased as per the label distribution, the larger dataset allows for training better models.
\end{abstract}

\section{Introduction}
\label{introduction}

\begin{table*}
\centering\small
\setlength{\tabcolsep}{3pt}
\begin{tabular}{@{}l@{\hspace{10\tabcolsep}}c@{\hspace{5\tabcolsep}}cccc@{\hspace{5\tabcolsep}}cccc@{}}
\toprule
\bf Argument source            & \bf Year & \multicolumn{4}{c@{\hspace{5\tabcolsep}}}{\bf Arguments}                                 & \multicolumn{4}{c@{}}{\bf Unique conclusions} \\
\cmidrule(r{5\tabcolsep}){3-6}
\cmidrule{7-10}
                               &          & \bf Train      & \bf Validation  & \bf Test       & $\sum$         & \bf Train      & \bf Validation  & \bf Test       & $\sum$         \\
\midrule
\it Main dataset \\                  
IBM-ArgQ-Rank-30kArgs          &  2019--20 & \phantom{}4576 & \phantom{}1526 & \phantom{}1266 & \phantom{}7368 & \phantom{00}46 & \phantom{00}15 & \phantom{00}10 & \phantom{00}71 \\
Conf.\ on the Future of Europe &  2021--22 & \phantom{0}591 & \phantom{0}280 & \phantom{0}227 & \phantom{}1098 & \phantom{0}232 & \phantom{0}119 & \phantom{00}80 & \phantom{0}431 \\
Group Discussion Ideas         &  2021--22 & \phantom{0}226 & \phantom{00}90 & \phantom{00}83 & \phantom{0}399 & \phantom{00}54 & \phantom{00}23 & \phantom{00}16 & \phantom{00}93 \\
\addlinespace
$\sum$ (main)                  &           & \phantom{}5393 & \phantom{}1896 & \phantom{}1576 & \phantom{}8865 & \phantom{0}332 & \phantom{0}157 & \phantom{0}106 & \phantom{0}595 \\
\midrule
\it Supplementary dataset  \\
Zhihu                          &      2021 &              - & \phantom{0}100 &              - & \phantom{0}100 &              - & \phantom{00}12 &              - & \phantom{00}12 \\
Nahj al-Balagha                & 900--1000 &              - &              - & \phantom{0}279 & \phantom{0}279 &              - &              - & \phantom{00}81 & \phantom{00}81 \\
The New York Times             &  2020--21 &              - &              - & \phantom{00}80 & \phantom{00}80 &              - &              - & \phantom{00}80 & \phantom{00}80 \\
\addlinespace
 $\sum$ (supplementary)        &           &              - & \phantom{0}100 & \phantom{0}359 & \phantom{0}459 &              - & \phantom{00}12 & \phantom{0}161 & \phantom{0}173 \\
\midrule
$\sum$ (complete)              &           & \phantom{}5393 & \phantom{}1996 & \phantom{}1935 & \phantom{}9324 & \phantom{0}332 & \phantom{0}169 & \phantom{0}267 & \phantom{0}768 \\
\bottomrule        
\end{tabular}
\caption{Key statistics of the main and supplementary dataset by argument source. Additional 1047 arguments have been collected from religious sources, but are excluded here as they have not been annotated yet (cf.\ Section~\ref{nahj-al-balagha}).}
\label{table-argument-values-dataset-overview}
\end{table*}

Why might one person find an argument more persuasive than someone else? One answer to this question is rooted in the values they hold. Although people might share a set of values, the priority they give to these values can be different (e.g. should \textit{having privacy} be considered more important than \textit{having a safe country?}). Such differences in priority can prevent people from finding common ground on a debatable topic or cause even more dispute. Moreover, differences in value priorities exist not only between individuals but also between cultures, which can cause disagreements. 

\begin{figure}
\includegraphics[scale=.75]{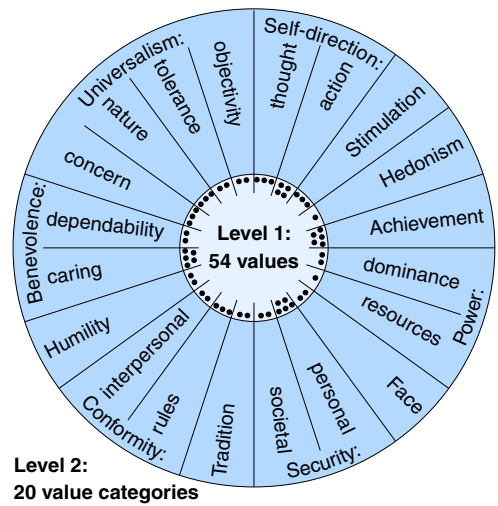}
\caption{The employed value taxonomy of 20~value categories and their associated 54~values (shown as black dots), the levels~2 and~1 from \citet{kiesel:2022b}. Categories that tend to conflict are placed on opposite sites. Illustration adapted from \cite{schwartz:1994}}
\label{schwartz-value-continuum-circle}
\end{figure}

Within computational linguistics, human values can provide context to categorize, compare, and evaluate argumentative statements, allowing for several applications: to inform social science research on values through large-scale datasets; to assess argumentation; to generate or select arguments for a target audience; and to identify opposing and shared values on both sides of a controversial topic. Probably the most widespread value categorization used in NLP is that of \citet{schwartz:1994}, shown (adapted) in Figure~\ref{schwartz-value-continuum-circle}, and used in the paper at hand. 

In order to tackle the challenges of human value identification---such as the wide variety of values, their often implicit use, and their ambiguous definition---we previously developed the practical foundations for AI-based identification systems \cite{kiesel:2022b}: a consolidated multi-level taxonomy based on extensive taxonomization by social scientists and an annotated dataset of 5\,270~arguments, the Webis-ArgValues-22. However, the existing dataset has two main shortcomings:
(i)~it is comparably small for training or tuning a machine learning model that needs to capture the (yet unknown) linguistic features that identify each human value;
(ii)~95\% of its arguments stem from a single background (the USA), thus hindering the development of cross-cultural value detection models.

% (https://aclanthology.org/2022.findings-acl.20.pdf)

In this work, we aim to fill these gaps for the automatic human value identification task by proposing an extension to the existing dataset: the Touch{\'e}23-ValueEval. It contains 9\,324 arguments on a variety of statements written in different styles, including religious texts (Nahj al-Balagha), political discussions (Group Discussion Ideas), free-text arguments (IBM-ArgQ-Rank-30kArgs), newspaper articles (The New York Times), community discussions (Zhihu), and democratic discourse (Conference on the Future of Europe). Moreover, we broaden the variety of arguments in terms of represented cultures and territories, as well as in terms of historical perspective. The proposed dataset was collected and annotated for the SemEval 2023 Task 4. ValueEval: Identification of Human Values behind Arguments%
\footnote{\scriptsize\url{https://touche.webis.de/semeval23/touche23-web}}
and is publicly available online.%
\footnote{Dataset: \scriptsize\url{https://doi.org/10.5281/zenodo.6814563}}

\section{Collecting Arguments}
\label{dataset}

\begin{table*}
\small\setlength{\tabcolsep}{6pt}
\centering
\begin{tabular}{@{}r@{\hspace{4pt}}p{10.15cm}p{2.8cm}>{\raggedright\arraybackslash}p{1.9cm}@{}}
\toprule
\multicolumn{2}{@{}l}{\bf Argument} & \bf Value categories & \bf Source \\
\midrule
% A30361
$\circ$ &
Con ``We should end the use of economic sanctions'':\newline
Economic sanctions provide security and ensure that citizens are treated fairly.
&
Security: societal,\newline % Have a safe country,\newline
Universalism: concern % Be just
& IBM-ArgQ-Rank-30kArgs \\
\addlinespace
% E07182
$\circ$ &
Pro ``We need a better migration policy.'':\newline
Discussing what happened in the past between Africa and Europe is useless. All slaves and their owners died a long time ago. You cannot blame the grandchildren.
&
Universalism: concern % Be just
& Conf. on the Future of Europe \\
\addlinespace
% D27068
$\circ$ &
Con ``Rapists should be tortured'':\newline
Throughout India, many false rape cases are being registered these days. Torturing all of the accused persons causes torture to innocent persons too.
&
Security: societal,\newline % Have a safe country,\newline Have a stable society,\newline
Universalism: concern % Be just
& Group Discussion Ideas \\
\addlinespace
% F01267
$\circ$ &
Con ``We should secretly give our help to the poor'':\newline
By showing others how to help the poor, we spread this work in the society.
&
Benevolence: caring,\newline % Be helpful,\newline
Universalism: concern % Be just
& Nahj al-Balagha \\
\addlinespace
% C26071
$\circ$ &
Con ``We should crack down on unreasonably high incomes.'':\newline
If the key to an individual's standard of living does not lie in income, then it is useless to simply regulate income.
&
Security: personal,\newline % Have a comfortable life,\newline
Universalism: concern% Be just
& Zhihu\\
\addlinespace
% G01042
$\circ$ &
Pro ``All of this is a sharp departure from a long history of judicial solicitude toward state powers during epidemics.'':\newline
In the past, when epidemics have threatened white Americans and those with political clout, courts found ways to uphold broad state powers.
&
Power: dominance,\newline % Have the right to command,\newline
Universalism: concern % Be just
& The New York Times\\
\bottomrule
\end{tabular}
\caption{Six example arguments (stance, conclusion, and premise) and their annotated value categories. We selected these to showcase different ways for resorting to {\em be just}, which is a value of the category {\em Universalism: concern}.}
\label{table-argument-values-examples}
\end{table*}

To investigate approaches for the automated detection of human values behind arguments, we collected a dataset of 9324~arguments. As in our previous publication on human value detection~\cite{kiesel:2022b}, each argument consists of one premise, one conclusion, and a stance attribute indicating whether the premise is in favor of (pro) or against (con) the conclusion. About half of the arguments (4\,569; 49\%) are taken from the existing Webis-ArgValues-22 dataset~\cite{kiesel:2022b}. The other half comprises new arguments, partially taken from the same sources as the Webis-ArgValues-22 (3\,298; 69\%), with the remaining arguments being from entirely new sources (1\,457; 31\%).

% NOTES (new arguments for old sources)
% IBM: 2999
% GD:   299

Table~\ref{table-argument-values-dataset-overview} provides key figures for the data, both for the main dataset used for the main ValueEval'23 leaderboard and for the supplementary dataset used for checking the robustness of approaches.

For the main leaderboard, we provide the main dataset as three separate sets as it is customary in machine-learning tasks, namely one set each for training, validation, and testing. The main dataset is compiled of arguments from three sources (see below), with approximately the same distribution in training, validation, and testing. To avoid train-test leakage from argument similarity, we ensured that all arguments with the same conclusions (but different premises) were in the same set. The ground-truth for the test dataset has been kept secret from participants for the duration of the ValueEval'23 competition. 

In addition to the main dataset, we collected a supplementary dataset of arguments that are quite different from the ones in the main dataset in terms of both written form and ethical reasoning. We kept this dataset separate from the main dataset to evaluate model performance both in the same setting as it was trained on and, as a challenge of generalizability, in a different setting.

The following sections describe for each source the source itself, our collection process, and our preprocessing of the arguments. For illustration, Table~\ref{table-argument-values-examples} provides one example argument per source.

%%%%%%%%%%%%%%%%%%%%%%%%%%%%%%%%%%%%%%%%%%%%%%%%%%%%%%%%%%%%%%%%%%%%%%%%%%%%%%%%
\subsection{IBM-ArgQ-Rank-30kArgs}
The original Webis-ArgValues-22 dataset contains 5\,020~arguments from the IBM-ArgQ-Rank-30kArgs dataset~\cite{gretz:2020}. We expand the dataset by including 2\,999~more arguments from this source. However, to avoid train-test leakage as mentioned above, we also had to exclude 651~arguments of the Webis-ArgValues-22 for which the conclusion is contained in the new test set.
\paragraph{Source}
For the IBM~dataset, crowdworkers were tasked to write one supporting and one contesting argument for one of 71~common controversial topics. The dataset totals 30\,497~arguments, each of which is annotated by crowdworkers for quality. The employed notion of high quality is: ``if a person preparing a speech on the topic will be likely to use the argument as is in [their] speech.'' \cite{gretz:2020}
\paragraph{Collection process}
We adopted the process that we used for the Webis-ArgValues-22: We sampled from the IBM~dataset only arguments where at least half of crowdworkers agreed that they are of high quality. We used the topics as conclusions and the ``arguments'' as respective premises.
\paragraph{Preprocessing}
We also adopted the same preprocessing approach: We manually corrected encoding errors in the text body of each argument, ensured a uniform character set for punctuation, and formatted arguments to be HTML compatible.

%%%%%%%%%%%%%%%%%%%%%%%%%%%%%%%%%%%%%%%%%%%%%%%%%%%%%%%%%%%%%%%%%%%%%%%%%%%%%%%%

\subsection{Conference on the Future of Europe}
\label{cofe}

The CoFE subpart consists of 1\,098 arguments for 431 unique conclusions, collected from the Conference on the Future of Europe portal.%
\footnote{\scriptsize\url{https://futureu.europa.eu}}
  
\paragraph{Source}
Conference on the Future of Europe was an online participatory democracy platform intended to involve citizens, experts and EU institutions in a dialogue focused on the future direction and legitimacy of Europe. CoFE was designed as a user-led series of debates, where anyone could give a proposal in any of the EU24 languages. For each of the proposals, any other user could endorse or criticize the proposals (similar to a like button), comment on them or reply to other comments. 

\paragraph{Collection Process}
In our work, we used the CoFE dataset \cite{barriere-etal-2022-cofe}, which contains more than 20 thousand comments on around 4.2 thousand proposals in 26 languages. English, German, and French are the main languages of the platform.  All the texts are automatically translated into any of the EU24 languages. A subset of the comments  in the dataset ($\approx$35\%) was labelled by users themselves, expressing their stance towards the proposition, around 6\% was annotated by experts, while the rest of the comments remain unlabeled.

\paragraph{Preprocessing}
Due to the limited time available, we focused on the proposals originally written in English. Out of 6\,985 available comment/proposal pairs containing user-annotations in the CoFE dataset, we preprocessed 1\,098 comments coming from 431 debates. We manually identified a conclusion in each of the proposals and one or more premises in the corresponding comments. We manually ensured that the resulting arguments had a similar length and structure to those in the Webis-ArgValues-22 dataset.

%%%%%%%%%%%%%%%%%%%%%%%%%%%%%%%%%%%%%%%%%%%%%%%%%%%%%%%%%%%%%%%%%%%%%%%%%%%%%%%%
\subsection{Group Discussion Ideas}
\label{group-discussion-ideas}
We extended the 100~arguments of the ``India'' part of the Webis-ArgValues-22, collected from the Group Discussion Ideas web page%
\footnote{\scriptsize\url{https://www.groupdiscussionideas.com}} by including 299~new arguments from the same source.
% In addition to the 100~arguments from the Group Discussion Ideas web page%
% \footnote{\scriptsize\url{https://www.groupdiscussionideas.com}}
% that are part of the Webis-ArgValues-22, we collected 299~new arguments.
\paragraph{Source}
This web page collects pros and cons on various topics covered in Indian news to help users support discussions in English. As the web page says, its goal is ``to provide all the valid points for the trending topics, so that the readers will be equipped with the required knowledge'' for a group discussion or debate. The web page currently lists a team of 16~authors. We received permission to distribute the arguments.
\paragraph{Collection process}
We crawled the web page and semi-automatically extracted arguments. For the original 100~arguments, we used a section of the web page called ``controversial debate topics 2021.'' For the additional 299~arguments, we extended our scope to include all topics from 2022.
\paragraph{Preprocessing}
We manually ensured that the arguments had a similar structure to those in the Webis-ArgValues-22 dataset by rewording and shortening them slightly if necessary.

%%%%%%%%%%%%%%%%%%%%%%%%%%%%%%%%%%%%%%%%%%%%%%%%%%%%%%%%%%%%%%%%%%%%%%%%%%%%%%%%
\subsection{Zhihu}
\label{zhihu}
We used the 100~arguments that were already part of the Webis-ArgValues-22 as-is. % rewrite?
These had been manually paraphrased from the recommendation and hotlist section of this Chinese question-answering website\footnote{\scriptsize\url{https://www.zhihu.com/explore}} and then manually translated into English. 

%%%%%%%%%%%%%%%%%%%%%%%%%%%%%%%%%%%%%%%%%%%%%%%%%%%%%%%%%%%%%%%%%%%%%%%%%%%%%%%%
\subsection{Nahj al-Balagha}
\label{nahj-al-balagha}
We collected and annotated 279~arguments from the Nahj al-Balagha, a collection of Islamic religious texts. These arguments are part of a larger dataset of 1\,326~arguments we collected from two Islamic sources, featuring advice and arguments on moral behavior. The remaining 1\,047~arguments have not been annotated yet due to time constraints. 

\paragraph{Source}
The books Nahj al-Balagha and Ghurar al-Hikam wa Durar al-Kalim contain moral aphorisms and eloquent content attributed to Ali ibn Abi Talib (600 CE, though published centuries later), who is known as one of the main Islamic elders. The Nahj al-Balagha includes more than 200~sermons, 80~letters, and 500~sayings. The Ghurar al-Hikam wa Durar al-Kalim contains 11\,000~pietistic and ethical short sayings. The two books were originally written in Arabic and have been subsequently translated into different languages. We employ standard translations of the books into Farsi.

\paragraph{Collection process}
We first manually extracted 302~premises from the Nahj al-Balagha: 181~were extracted verbatim and 121~were distilled from the text. The conclusions were deduced manually, with similar conclusions being unified. To balance the stance distribution, a few of the distilled premises were rephrased so that they are against the conclusion. The 279~annotated arguments are all taken from this set of 302~arguments; 23~unclear arguments were omitted from the annotation.

To enlarge the dataset for future uses, we implemented a semi-automated extraction pipeline, which we use to extract additional~1\,047 arguments from the texts. 878 of these were collected from Ghurar al-Hikam wa Durar al-Kalim, while the rest come from Nahj al-Balagha. We finetuned a pre-trained Persian BERT \cite{farahani:2021} language model over the extracted arguments and used it to identify potential further arguments, which were then checked and extracted like the ones mentioned above.

\newsavebox{\levelsafebox}
\newenvironment{level1box}
  {\newcommand\colorboxcolor{level1color}%
   \begin{lrbox}{\levelsafebox}%
   \begin{minipage}[t][2ex]{\widthof{Have the wisdom to accept others}}}
  {\end{minipage}\end{lrbox}%
   \setlength{\fboxsep}{0pt}\colorbox{\colorboxcolor}{\usebox{\levelsafebox}}}

\newcommand{\checkYes}{$\circ$}
\newcommand{\checkSWS}{{\color{gray}$\bullet$}}
\newcommand{\BarCell}[3][1]{%
  \setlength{\fboxsep}{0pt}%
    0.#3$\,$\fbox{\textcolor{#2}{\rule{#1\dimexpr 0.06pt*(#3)\relax}{5pt}}}}

\newcommand{\ColorBar}{gray}

\begin{table*}
\centering
\scriptsize%
%\fontsize{7}{8}\selectfont%
\setlength{\tabcolsep}{2pt}%

\begin{tabular}{@{}>{\columncolor{level2color}[0pt][1pt]}l>{\begin{level1box}}l<{\end{level1box}}@{\hspace{8pt}}lllllll}

\toprule
\multicolumn{2}{@{}l}{\bf Level} & \multicolumn{7}{c@{}}{\bf Dataset frequency (size; cf.\ Section~\ref{dataset})} \\
\cmidrule(r{8pt}){1-2}
\cmidrule{3-9}
\multicolumn{1}{@{}l}{\bf 2) Value category}
  & \multicolumn{1}{@{}l}{\bf 1) Value}               & \bf IBM (7368)           & \bf CoFE (1098)          & \bf GDI (399)            & \bf Zhihu (100)          & \bf Nahj (279)           & \bf NYT (80)             & \bf Total (9324) \\
\midrule

 Self-direction: thought & Be creative                & \BarCell{\ColorBar}{026} & \BarCell{\ColorBar}{025} & \BarCell{\ColorBar}{018} & \BarCell{\ColorBar}{040} & \BarCell{\ColorBar}{004} & \BarCell{\ColorBar}{000} & \BarCell{\ColorBar}{025} \\ 
 & Be curious                                         & \BarCell{\ColorBar}{045} & \BarCell{\ColorBar}{027} & \BarCell{\ColorBar}{045} & \BarCell{\ColorBar}{030} & \BarCell{\ColorBar}{004} & \BarCell{\ColorBar}{025} & \BarCell{\ColorBar}{041} \\  
 & Have freedom of thought                            & \BarCell{\ColorBar}{117} & \BarCell{\ColorBar}{054} & \BarCell{\ColorBar}{045} & \BarCell{\ColorBar}{000} & \BarCell{\ColorBar}{014} & \BarCell{\ColorBar}{000} & \BarCell{\ColorBar}{101} \\  
\addlinespace                                                                                                                                                     
 Self-direction: action & Be choosing own goals       & \BarCell{\ColorBar}{129} & \BarCell{\ColorBar}{105} & \BarCell{\ColorBar}{103} & \BarCell{\ColorBar}{030} & \BarCell{\ColorBar}{004} & \BarCell{\ColorBar}{000} & \BarCell{\ColorBar}{119} \\  
 & Be independent                                     & \BarCell{\ColorBar}{102} & \BarCell{\ColorBar}{109} & \BarCell{\ColorBar}{098} & \BarCell{\ColorBar}{030} & \BarCell{\ColorBar}{011} & \BarCell{\ColorBar}{000} & \BarCell{\ColorBar}{098} \\  
 & Have freedom of action                             & \BarCell{\ColorBar}{181} & \BarCell{\ColorBar}{120} & \BarCell{\ColorBar}{098} & \BarCell{\ColorBar}{030} & \BarCell{\ColorBar}{029} & \BarCell{\ColorBar}{000} & \BarCell{\ColorBar}{163} \\  
 & Have privacy                                       & \BarCell{\ColorBar}{017} & \BarCell{\ColorBar}{012} & \BarCell{\ColorBar}{063} & \BarCell{\ColorBar}{040} & \BarCell{\ColorBar}{004} & \BarCell{\ColorBar}{012} & \BarCell{\ColorBar}{018} \\  
\addlinespace                                                                                                                                                     
 Stimulation & Have an exciting life                  & \BarCell{\ColorBar}{017} & \BarCell{\ColorBar}{004} & \BarCell{\ColorBar}{018} & \BarCell{\ColorBar}{000} & \BarCell{\ColorBar}{000} & \BarCell{\ColorBar}{000} & \BarCell{\ColorBar}{015} \\  
 & Have a varied life                                 & \BarCell{\ColorBar}{038} & \BarCell{\ColorBar}{027} & \BarCell{\ColorBar}{040} & \BarCell{\ColorBar}{000} & \BarCell{\ColorBar}{004} & \BarCell{\ColorBar}{000} & \BarCell{\ColorBar}{035} \\  
 & Be daring                                          & \BarCell{\ColorBar}{010} & \BarCell{\ColorBar}{007} & \BarCell{\ColorBar}{000} & \BarCell{\ColorBar}{000} & \BarCell{\ColorBar}{004} & \BarCell{\ColorBar}{000} & \BarCell{\ColorBar}{009} \\  
\addlinespace                                                                                                                                                     
 Hedonism & Have pleasure                             & \BarCell{\ColorBar}{038} & \BarCell{\ColorBar}{005} & \BarCell{\ColorBar}{040} & \BarCell{\ColorBar}{020} & \BarCell{\ColorBar}{014} & \BarCell{\ColorBar}{012} & \BarCell{\ColorBar}{033} \\  
\addlinespace                                                                                                                                                     
 Achievement & Be ambitious                           & \BarCell{\ColorBar}{042} & \BarCell{\ColorBar}{046} & \BarCell{\ColorBar}{068} & \BarCell{\ColorBar}{050} & \BarCell{\ColorBar}{047} & \BarCell{\ColorBar}{000} & \BarCell{\ColorBar}{043} \\  
 & Have success                                       & \BarCell{\ColorBar}{120} & \BarCell{\ColorBar}{097} & \BarCell{\ColorBar}{148} & \BarCell{\ColorBar}{160} & \BarCell{\ColorBar}{068} & \BarCell{\ColorBar}{012} & \BarCell{\ColorBar}{116} \\  
 & Be capable                                         & \BarCell{\ColorBar}{159} & \BarCell{\ColorBar}{215} & \BarCell{\ColorBar}{253} & \BarCell{\ColorBar}{200} & \BarCell{\ColorBar}{068} & \BarCell{\ColorBar}{100} & \BarCell{\ColorBar}{167} \\  
 & Be intellectual                                    & \BarCell{\ColorBar}{067} & \BarCell{\ColorBar}{040} & \BarCell{\ColorBar}{080} & \BarCell{\ColorBar}{130} & \BarCell{\ColorBar}{097} & \BarCell{\ColorBar}{062} & \BarCell{\ColorBar}{066} \\  
 & Be courageous                                      & \BarCell{\ColorBar}{010} & \BarCell{\ColorBar}{008} & \BarCell{\ColorBar}{003} & \BarCell{\ColorBar}{000} & \BarCell{\ColorBar}{022} & \BarCell{\ColorBar}{012} & \BarCell{\ColorBar}{009} \\  
\addlinespace                                                                                                                                                     
 Power: dominance & Have influence                    & \BarCell{\ColorBar}{057} & \BarCell{\ColorBar}{101} & \BarCell{\ColorBar}{088} & \BarCell{\ColorBar}{010} & \BarCell{\ColorBar}{011} & \BarCell{\ColorBar}{000} & \BarCell{\ColorBar}{061} \\  
 & Have the right to command                          & \BarCell{\ColorBar}{037} & \BarCell{\ColorBar}{100} & \BarCell{\ColorBar}{045} & \BarCell{\ColorBar}{000} & \BarCell{\ColorBar}{007} & \BarCell{\ColorBar}{012} & \BarCell{\ColorBar}{043} \\  
\addlinespace                                                                                                                                                     
 Power: resources & Have wealth                       & \BarCell{\ColorBar}{099} & \BarCell{\ColorBar}{084} & \BarCell{\ColorBar}{100} & \BarCell{\ColorBar}{190} & \BarCell{\ColorBar}{014} & \BarCell{\ColorBar}{000} & \BarCell{\ColorBar}{095} \\  
\addlinespace                                                                                                                                                     
 Face & Have social recognition                       & \BarCell{\ColorBar}{047} & \BarCell{\ColorBar}{055} & \BarCell{\ColorBar}{068} & \BarCell{\ColorBar}{000} & \BarCell{\ColorBar}{032} & \BarCell{\ColorBar}{000} & \BarCell{\ColorBar}{048} \\  
 & Have a good reputation                             & \BarCell{\ColorBar}{022} & \BarCell{\ColorBar}{040} & \BarCell{\ColorBar}{028} & \BarCell{\ColorBar}{010} & \BarCell{\ColorBar}{111} & \BarCell{\ColorBar}{025} & \BarCell{\ColorBar}{027} \\  
\addlinespace                                                                                                                                                     
 Security: personal & Have a sense of belonging       & \BarCell{\ColorBar}{077} & \BarCell{\ColorBar}{108} & \BarCell{\ColorBar}{075} & \BarCell{\ColorBar}{010} & \BarCell{\ColorBar}{075} & \BarCell{\ColorBar}{025} & \BarCell{\ColorBar}{080} \\  
 & Have good health                                   & \BarCell{\ColorBar}{136} & \BarCell{\ColorBar}{066} & \BarCell{\ColorBar}{125} & \BarCell{\ColorBar}{030} & \BarCell{\ColorBar}{036} & \BarCell{\ColorBar}{275} & \BarCell{\ColorBar}{124} \\  
 & Have no debts                                      & \BarCell{\ColorBar}{056} & \BarCell{\ColorBar}{061} & \BarCell{\ColorBar}{068} & \BarCell{\ColorBar}{020} & \BarCell{\ColorBar}{004} & \BarCell{\ColorBar}{000} & \BarCell{\ColorBar}{055} \\  
 & Be neat and tidy                                   & \BarCell{\ColorBar}{003} & \BarCell{\ColorBar}{006} & \BarCell{\ColorBar}{003} & \BarCell{\ColorBar}{000} & \BarCell{\ColorBar}{004} & \BarCell{\ColorBar}{000} & \BarCell{\ColorBar}{003} \\  
 & Have a comfortable life                            & \BarCell{\ColorBar}{185} & \BarCell{\ColorBar}{158} & \BarCell{\ColorBar}{251} & \BarCell{\ColorBar}{260} & \BarCell{\ColorBar}{129} & \BarCell{\ColorBar}{075} & \BarCell{\ColorBar}{183} \\  
\addlinespace                                                                                                                                                     
 Security: societal & Have a safe country             & \BarCell{\ColorBar}{185} & \BarCell{\ColorBar}{226} & \BarCell{\ColorBar}{160} & \BarCell{\ColorBar}{030} & \BarCell{\ColorBar}{007} & \BarCell{\ColorBar}{062} & \BarCell{\ColorBar}{180} \\  
 & Have a stable society                              & \BarCell{\ColorBar}{190} & \BarCell{\ColorBar}{237} & \BarCell{\ColorBar}{135} & \BarCell{\ColorBar}{300} & \BarCell{\ColorBar}{029} & \BarCell{\ColorBar}{075} & \BarCell{\ColorBar}{189} \\  
\addlinespace                                                                                                                                                     
 Tradition & Be respecting traditions                 & \BarCell{\ColorBar}{077} & \BarCell{\ColorBar}{105} & \BarCell{\ColorBar}{040} & \BarCell{\ColorBar}{000} & \BarCell{\ColorBar}{000} & \BarCell{\ColorBar}{000} & \BarCell{\ColorBar}{075} \\  
 & Be holding religious faith                         & \BarCell{\ColorBar}{046} & \BarCell{\ColorBar}{008} & \BarCell{\ColorBar}{023} & \BarCell{\ColorBar}{000} & \BarCell{\ColorBar}{100} & \BarCell{\ColorBar}{000} & \BarCell{\ColorBar}{041} \\  
\addlinespace                                                                                                                                                     
 Conformity: rules & Be compliant                     & \BarCell{\ColorBar}{124} & \BarCell{\ColorBar}{179} & \BarCell{\ColorBar}{120} & \BarCell{\ColorBar}{070} & \BarCell{\ColorBar}{022} & \BarCell{\ColorBar}{000} & \BarCell{\ColorBar}{126} \\  
 & Be self-disciplined                                & \BarCell{\ColorBar}{028} & \BarCell{\ColorBar}{016} & \BarCell{\ColorBar}{020} & \BarCell{\ColorBar}{030} & \BarCell{\ColorBar}{025} & \BarCell{\ColorBar}{012} & \BarCell{\ColorBar}{026} \\  
 & Be behaving properly                               & \BarCell{\ColorBar}{125} & \BarCell{\ColorBar}{061} & \BarCell{\ColorBar}{095} & \BarCell{\ColorBar}{070} & \BarCell{\ColorBar}{043} & \BarCell{\ColorBar}{038} & \BarCell{\ColorBar}{113} \\  
\addlinespace                                                                                                                                                     
 Conformity: interpersonal & Be polite                & \BarCell{\ColorBar}{031} & \BarCell{\ColorBar}{009} & \BarCell{\ColorBar}{023} & \BarCell{\ColorBar}{010} & \BarCell{\ColorBar}{029} & \BarCell{\ColorBar}{000} & \BarCell{\ColorBar}{027} \\  
 & Be honoring elders                                 & \BarCell{\ColorBar}{010} & \BarCell{\ColorBar}{003} & \BarCell{\ColorBar}{010} & \BarCell{\ColorBar}{000} & \BarCell{\ColorBar}{004} & \BarCell{\ColorBar}{012} & \BarCell{\ColorBar}{009} \\  
\addlinespace                                                                                                                                                     
 Humility & Be humble                                 & \BarCell{\ColorBar}{012} & \BarCell{\ColorBar}{010} & \BarCell{\ColorBar}{005} & \BarCell{\ColorBar}{020} & \BarCell{\ColorBar}{043} & \BarCell{\ColorBar}{038} & \BarCell{\ColorBar}{013} \\  
 & Have life accepted as is                           & \BarCell{\ColorBar}{066} & \BarCell{\ColorBar}{031} & \BarCell{\ColorBar}{018} & \BarCell{\ColorBar}{040} & \BarCell{\ColorBar}{036} & \BarCell{\ColorBar}{025} & \BarCell{\ColorBar}{058} \\  
\addlinespace                                                                                                                                                     
 Benevolence: caring & Be helpful                     & \BarCell{\ColorBar}{139} & \BarCell{\ColorBar}{122} & \BarCell{\ColorBar}{133} & \BarCell{\ColorBar}{030} & \BarCell{\ColorBar}{039} & \BarCell{\ColorBar}{038} & \BarCell{\ColorBar}{132} \\  
 & Be honest                                          & \BarCell{\ColorBar}{043} & \BarCell{\ColorBar}{046} & \BarCell{\ColorBar}{060} & \BarCell{\ColorBar}{010} & \BarCell{\ColorBar}{014} & \BarCell{\ColorBar}{012} & \BarCell{\ColorBar}{043} \\  
 & Be forgiving                                       & \BarCell{\ColorBar}{018} & \BarCell{\ColorBar}{005} & \BarCell{\ColorBar}{005} & \BarCell{\ColorBar}{000} & \BarCell{\ColorBar}{007} & \BarCell{\ColorBar}{000} & \BarCell{\ColorBar}{015} \\  
 & Have the own family secured                        & \BarCell{\ColorBar}{074} & \BarCell{\ColorBar}{030} & \BarCell{\ColorBar}{038} & \BarCell{\ColorBar}{090} & \BarCell{\ColorBar}{004} & \BarCell{\ColorBar}{000} & \BarCell{\ColorBar}{065} \\  
 & Be loving                                          & \BarCell{\ColorBar}{045} & \BarCell{\ColorBar}{010} & \BarCell{\ColorBar}{060} & \BarCell{\ColorBar}{020} & \BarCell{\ColorBar}{032} & \BarCell{\ColorBar}{012} & \BarCell{\ColorBar}{041} \\  
\addlinespace                                                                                                                                                     
 Benevolence: dependability & Be responsible          & \BarCell{\ColorBar}{128} & \BarCell{\ColorBar}{189} & \BarCell{\ColorBar}{143} & \BarCell{\ColorBar}{030} & \BarCell{\ColorBar}{047} & \BarCell{\ColorBar}{150} & \BarCell{\ColorBar}{132} \\  
 & Have loyalty towards friends                       & \BarCell{\ColorBar}{004} & \BarCell{\ColorBar}{002} & \BarCell{\ColorBar}{008} & \BarCell{\ColorBar}{000} & \BarCell{\ColorBar}{018} & \BarCell{\ColorBar}{000} & \BarCell{\ColorBar}{004} \\  
\addlinespace                                                                                                                                                     
 Universalism: concern & Have equality                & \BarCell{\ColorBar}{168} & \BarCell{\ColorBar}{019} & \BarCell{\ColorBar}{216} & \BarCell{\ColorBar}{090} & \BarCell{\ColorBar}{011} & \BarCell{\ColorBar}{088} & \BarCell{\ColorBar}{167} \\  
 & Be just                                            & \BarCell{\ColorBar}{252} & \BarCell{\ColorBar}{232} & \BarCell{\ColorBar}{221} & \BarCell{\ColorBar}{180} & \BarCell{\ColorBar}{025} & \BarCell{\ColorBar}{100} & \BarCell{\ColorBar}{240} \\  
 & Have a world at peace                              & \BarCell{\ColorBar}{077} & \BarCell{\ColorBar}{084} & \BarCell{\ColorBar}{030} & \BarCell{\ColorBar}{000} & \BarCell{\ColorBar}{029} & \BarCell{\ColorBar}{012} & \BarCell{\ColorBar}{073} \\  
\addlinespace                                                                                                                                                     
 Universalism: nature & Be protecting the environment & \BarCell{\ColorBar}{036} & \BarCell{\ColorBar}{156} & \BarCell{\ColorBar}{055} & \BarCell{\ColorBar}{080} & \BarCell{\ColorBar}{000} & \BarCell{\ColorBar}{000} & \BarCell{\ColorBar}{050} \\  
 & Have harmony with nature                           & \BarCell{\ColorBar}{052} & \BarCell{\ColorBar}{099} & \BarCell{\ColorBar}{065} & \BarCell{\ColorBar}{050} & \BarCell{\ColorBar}{004} & \BarCell{\ColorBar}{012} & \BarCell{\ColorBar}{057} \\  
 & Have a world of beauty                             & \BarCell{\ColorBar}{012} & \BarCell{\ColorBar}{005} & \BarCell{\ColorBar}{000} & \BarCell{\ColorBar}{000} & \BarCell{\ColorBar}{004} & \BarCell{\ColorBar}{000} & \BarCell{\ColorBar}{010} \\  
\addlinespace                                                                                                                                                     
 Universalism: tolerance & Be broadminded             & \BarCell{\ColorBar}{094} & \BarCell{\ColorBar}{069} & \BarCell{\ColorBar}{080} & \BarCell{\ColorBar}{010} & \BarCell{\ColorBar}{014} & \BarCell{\ColorBar}{012} & \BarCell{\ColorBar}{086} \\  
 & Have the wisdom to accept others                   & \BarCell{\ColorBar}{053} & \BarCell{\ColorBar}{069} & \BarCell{\ColorBar}{033} & \BarCell{\ColorBar}{010} & \BarCell{\ColorBar}{007} & \BarCell{\ColorBar}{000} & \BarCell{\ColorBar}{052} \\  
\addlinespace                                                                                                                                                     
 Universalism: objectivity & Be logical               & \BarCell{\ColorBar}{101} & \BarCell{\ColorBar}{210} & \BarCell{\ColorBar}{193} & \BarCell{\ColorBar}{120} & \BarCell{\ColorBar}{011} & \BarCell{\ColorBar}{125} & \BarCell{\ColorBar}{115} \\  
 & Have an objective view                             & \BarCell{\ColorBar}{127} & \BarCell{\ColorBar}{172} & \BarCell{\ColorBar}{163} & \BarCell{\ColorBar}{160} & \BarCell{\ColorBar}{065} & \BarCell{\ColorBar}{150} & \BarCell{\ColorBar}{133} \\  

\bottomrule
\end{tabular}

\caption{The 54~values of the taxonomy and dataset frequency per source: IBM-ArgQ-Rank-30kArgs (IBM), Conference on the Future of Europe (CoFE), Group Discussion Ideas (GDI), Zhihu, Nahj al-Balagha (Nahj), and The New York Times (NYT), as well as overall dataset frequency.} 
\label{table-values-description}

\end{table*}

\paragraph{Preprocessing}
We manually translated the arguments into English and had another annotator check the whole dataset to remove ambiguous arguments. 

%%%%%%%%%%%%%%%%%%%%%%%%%%%%%%%%%%%%%%%%%%%%%%%%%%%%%%%%%%%%%%%%%%%%%%%%%%%%%%%%
\subsection{The New York Times}
\label{nyt}
We collected 80~arguments from news articles published in The New York Times.%
\footnote{\scriptsize\url{https://www.nytimes.com}}
At the time of writing, we are in the process of obtaining permission to publish the arguments. Until then, we provide Python software that extracts the arguments from the Internet Archive.%
\footnote{\hbadness=10000{\scriptsize\url{https://github.com/touche-webis-de/touche-code/tree/main/semeval23/human-value-detection/nyt-downloader}}}

\paragraph{Source}
The New York Times is a renowned US-American daily newspaper that is available in print and via an online subscription.

\paragraph{Collection process}
We selected 12~editorials, published between July~2020 and May~2021, with at least one of the New York Times keywords \textit{coronavirus (2019-ncov)}, \textit{vaccination and immunization}, and \textit{epidemics}. We manually selected texts with an overall high quality of argumentation, as assessed by three linguistically trained annotators. 

\paragraph{Preprocessing}
The premises, conclusions, and stances were manually annotated by four annotators (three per text), and these annotations were curated by two linguist experts. The test set does not comprise all arguments identified in the twelve texts, but rather a selection of especially clear ones, as established by the curators.

% NOTE
% For all purposes, the output of the script is for now what is in the dataset. The following is thus not needed
%
%%% Furthermore, for 63 of the 80 arguments, \bscom[NH]{the conclusion comes before the premise in the original text, as opposed to the output of the above-mentioned script, where all arguments have been standardized for the premise to come before the conclusion}{the original text mentions the conclusion before the premise, as opposed to the output of the above-denoted script, where all arguments have been standardized for the premise to be mentioned before the conclusion} (without altering the linguistic material). \textcolor{red}{Computational linguists, please reformulate this last bit so that it makes sense to you.}

\section{Crowdsourcing the Annotation of Human Values behind Arguments}
\label{annotation}

We re-used the crowdsourcing setup of 3~human annotators per argument of \citet{kiesel:2022b} (Webis-ArgValues-22). For illustration, we reprint the screenshots of the annotation interface in Appendix~\ref{annotation-interface}. As the screenshots show, the interface contains annotation instructions (cf.\ Figure~\ref{argument-values-annotation-interface-top}) and uses yes/no questions for labeling each argument for each of the 54~level~1 values (cf.\ Figure~\ref{argument-values-annotation-interface-bottom}). Though the ValueEval'23 task uses only level~2 value categories, we kept the tried and tested annotation process both for consistency and to allow for approaches that work on level~1. We restricted annotation to the 27~annotators who passed the selection process for Webis-ArgValues-22, of which 13~returned to work under the same payment. In total, the annotators made 774\,360 yes/no annotations for 4\,780~new arguments. Like for Webis-ArgValues-22, we employed MACE \cite{hovy:2013} to fuse the annotations into a single ground truth. For quality assurance, we inspected all annotations for arguments from the Nahj al-Balagha and the New York Times, as well as those for which MACE's confidence was about 50:50. For this check, we analyzed 727~arguments, for which we changed the annotation if necessary. This check focused on the two supplementary test sets, as in these datasets the conclusion also often references values, which confused some crowdworkers.

\section{Analyzing the Dataset}
\label{analysis}

This section first presents an overview of the main statistics of our dataset, then highlights the similarities and differences among value distributions of the used sources. Finally, we report on the results of baseline experiments that investigate the influence of dataset extension on the task at hand.

\begin{figure}[t]
\centering\small
\includegraphics[width=\linewidth]{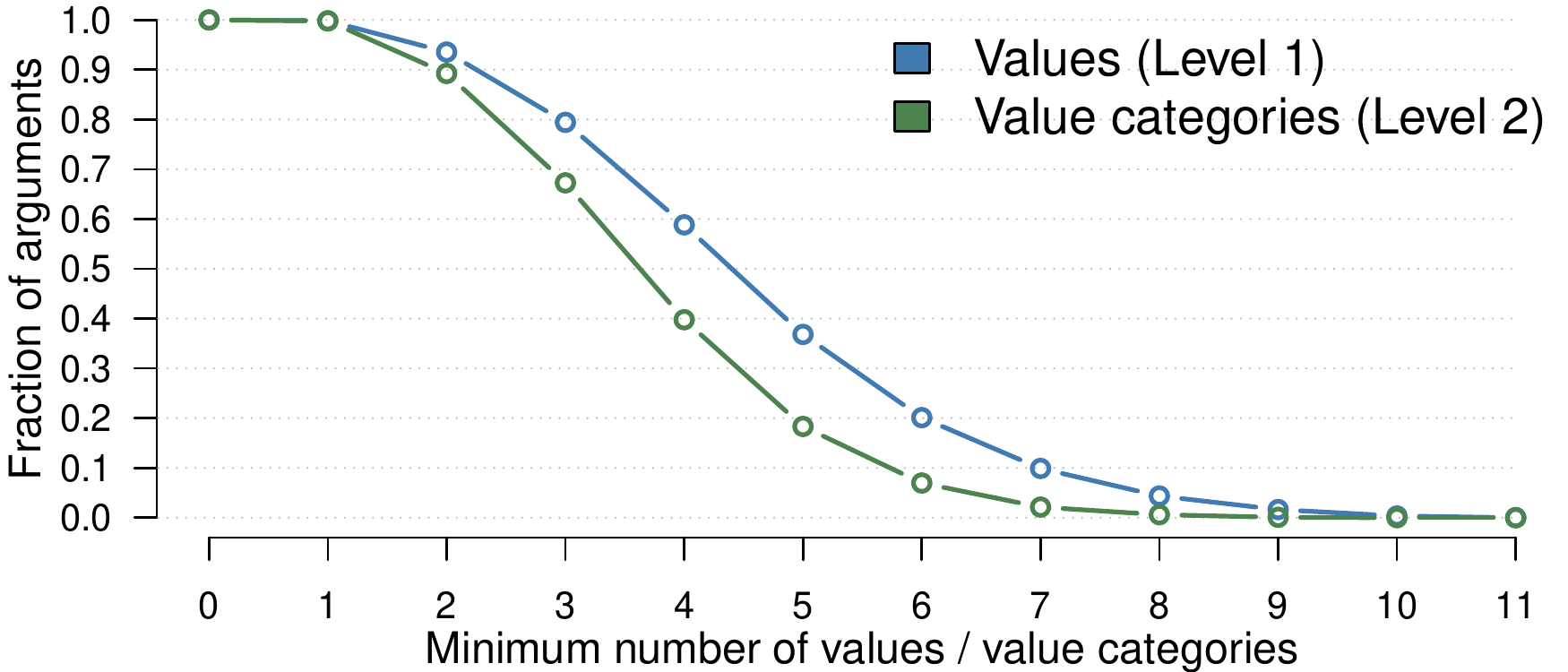}
\caption{Fraction of arguments in the complete dataset having a specific number of assigned values (out of~54) or value categories (out of~10) or more.}
\label{ecdf-number-of-labels}
\end{figure}

\paragraph{Overview statistics}
The dataset consists of 9\,324 unique premise-conclusion pairs. Each of the arguments is annotated for multiple values on two levels of granularity. As Figure~\ref{ecdf-number-of-labels} shows, 94\% of the arguments have at least 2 values, and 89\% have more than 2 value categories assigned to them. A total of 18 arguments (\textasciitilde{}0.19\%) have no assigned value to them (i.e., they resort to no ethical judgement).  The most frequent values in the dataset are \textit{Be just}, \textit{Have a stable society}, and \textit{Have a safe country}. More fine-grained distribution statistics for each of the values are shown in Table~\ref{table-values-description}.
The average length of a premise is 23.53 words, and that of a conclusion is 6.48 words. The stance distribution is generally balanced, with an approximate 10\% skew, however, towards the \textit{pro} label (cf.\ Table~\ref{table-mean-length}).

\begin{table}[t]
\footnotesize\centering%
\setlength{\tabcolsep}{2pt}%
\begin{tabular}{@{}l@{\hspace{2px}}cc@{\hspace{6pt}}cc@{}}
\toprule
% \bf Part & \multicolumn{2}{c}{\bf Conclusions} & \multicolumn{2}{c}{\bf Premises} & \multicolumn{2}{c}{\bf Stances}\\

                              & \multicolumn{2}{@{}c@{\hspace{6pt}}}{\bf Mean length} & \multicolumn{2}{@{}c@{}}{\bf Arguments} \\
\cmidrule(r{6pt}){2-3}
\cmidrule{4-5}
\bf Argument source           & \bf Concl.      & \bf Premise       & \bf Pro        & \bf Con        \\
\midrule
IBM-ArgQ-Rank-30kArgs         & \phantom{0}5.55 & 19.84             &           3824 &           3544 \\
Conf. on the Future of Europe &           11.35 & 39.59             & \phantom{0}750 & \phantom{0}348 \\
Group Discussion Ideas        & \phantom{0}7.87 & 45.27             & \phantom{0}250 & \phantom{0}149 \\
Zhihu                         & \phantom{0}8.19 & 27.51             & \phantom{00}59 & \phantom{00}41 \\
Nahj al-Balagha               & \phantom{0}5.58 & 22.40             & \phantom{0}224 & \phantom{00}55 \\
The New York Times            &           20.20 & 22.87             & \phantom{00}69 & \phantom{00}11 \\

\midrule[\cmidrulewidth]
$\sum$ (complete)             & \phantom{0}6.48 & 23.53             &           5176 &           4148 \\
\bottomrule
\end{tabular}
\caption{Mean length (number of space-separated tokens) in conclusions and premises and the stance distribution per source of the Touch{\'e}23-ValueEval dataset.}
\label{table-mean-length}
\end{table}

\begin{figure}
\centering
\includegraphics[scale=.45]{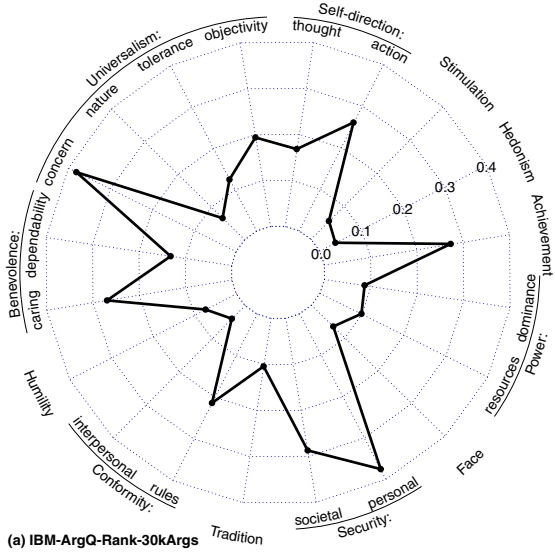}%
\hfill%
\includegraphics[scale=.45]{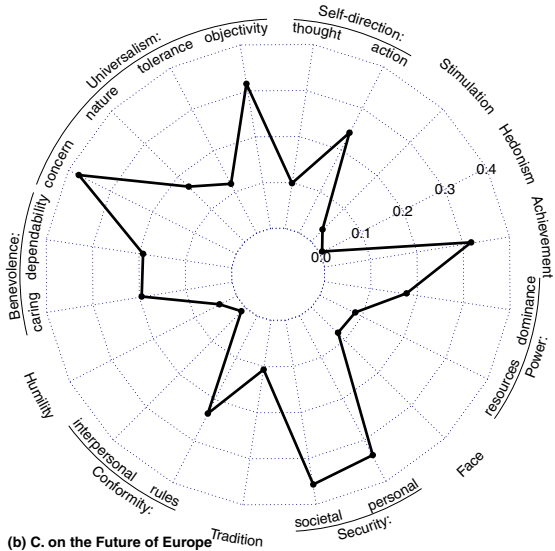}\\[3ex]%
\includegraphics[scale=.45]{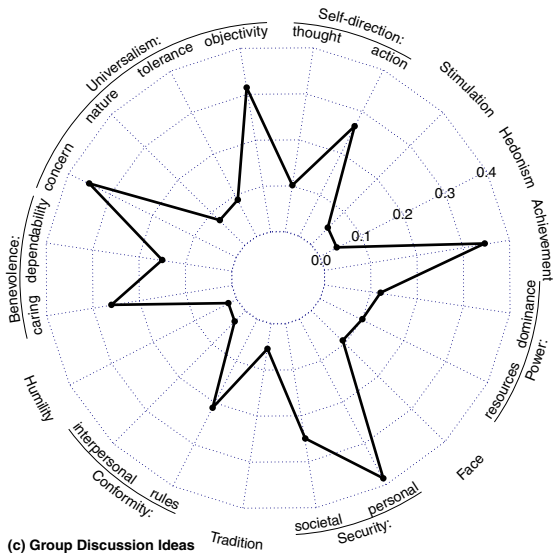}%
\caption{Distribution of value categories across the sources in the \textit{main} dataset.}
\label{level2-radars-main}
\end{figure}

\begin{figure*}
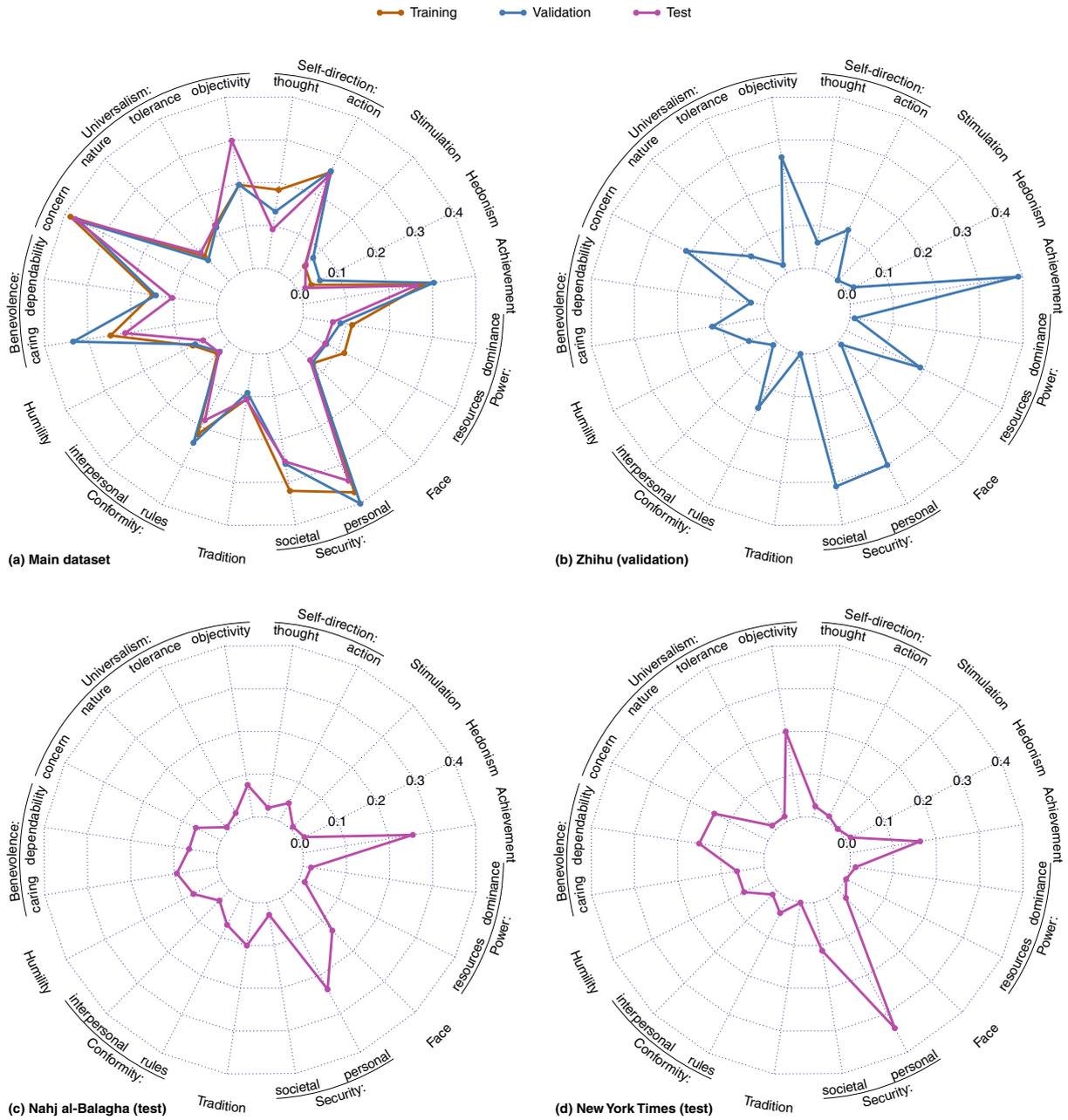

\centering
\includegraphics[scale=.48]{level2-radar-legend}\\[3ex]%
\includegraphics[scale=.48]{level2-radar-main}%
\hfill%
\includegraphics[scale=.48]{level2-radar-zhihu}\\[3ex]%
\includegraphics[scale=.48]{level2-radar-nahjalbalagha}
\hfill%
\includegraphics[scale=.48]{level2-radar-nyt}
\caption{Distribution of value categories across the training, validation and testing splits, as well as within the sources of the \textit{supplementary} dataset.}
\label{level2-radars}
\end{figure*}

\begin{table*}
\setlength{\tabcolsep}{3pt}\centering
\small
\begin{tabular}{@{}l@{\hspace{20pt}}cccc@{\hspace{10pt}}cccc@{\hspace{20pt}}cccc@{\hspace{10pt}}cccc@{}}
\toprule
\bf Model & \multicolumn{8}{@{}c@{\hspace{20pt}}}{\bf Values (Level 1)} & \multicolumn{8}{c@{}}{\bf Value categories (Level 2)} \\
\cmidrule(r{20pt}){2-9}
\cmidrule{10-17}
& \multicolumn{4}{@{}c@{\hspace{10pt}}}{\bf  Webis-ArgValues-22} & \multicolumn{4}{@{}c@{\hspace{20pt}}}{\bf Touch{\'e}23-ValueEval} & \multicolumn{4}{@{}c@{\hspace{10pt}}}{\bf  Webis-ArgValues-22} & \multicolumn{4}{@{}c}{\bf Touch{\'e}23-ValueEval} \\
\cmidrule(r{10pt}){2-5}
\cmidrule(r{20pt}){6-9}
\cmidrule(r{10pt}){10-13}
\cmidrule{14-17}
& \bf P    & \bf R    & \bf F$_1$& \bf Acc & \bf P    & \bf R    & \bf F$_1$& \bf Acc & \bf P    & \bf R    & \bf F$_1$& \bf Acc & \bf P    & \bf R    & \bf F$_1$& \bf Acc \\
\midrule
BERT
& 0.40     & \bf 0.19 & 0.25     & 0.92
& \bf 0.43 & \bf 0.19 & \bf 0.26 & \bf 0.94
& 0.39     & 0.30     & 0.34     & 0.84
& \bf 0.59 & \bf 0.35 & \bf 0.44 & \bf 0.88 \\
% SVM
% & 0.21     & \bf 0.19 & \bf 0.20 & 0.88
% & \bf 0.27 & 0.12     & 0.16     & \bf 0.92
% & 0.30     & \bf 0.30 & \bf 0.30 & 0.77
% & \bf 0.32 & 0.21     & 0.25     & \bf 0.83 \\
1-Baseline
& \bf 0.08 & \bf 1.00 & \bf 0.16 & \bf 0.08
& 0.07     & \bf 1.00 & 0.13     & 0.07
& \bf 0.18 & \bf 1.00 & \bf 0.28 & \bf 0.18
& 0.15     & \bf 1.00 & 0.26     & 0.15     \\
\bottomrule
\end{tabular}
\caption{Comparison of macro precision~(P), recall~(R), F$_1$-score~(F$_1$), and accuracy~(Acc) on respective test sets of Webis-ArgValues-22 and Touch{\'e}23-ValueEval by level.}
\label{table-compare-results}
\end{table*}

\paragraph{Value distributions}
Figures \ref{level2-radars-main} and \ref{level2-radars} depict the distribution of value categories (Level 2 in Figure \ref{schwartz-value-continuum-circle}) across the train/validation/test splits, as well as within each of the data sources.
As for the sources used in the \textit{main} dataset, Figure \ref{level2-radars-main} demonstrates that all three sources share similar value categories distribution with slight fluctuations. For instance, discussion boards (Group Discussion Ideas, Conference on the Future of Europe) seem to value \textit{Universalism: Objectivity} considerably more than respondents for IBM-ArgQ-Rank-30kArgs. Besides that, the most  common category for all three sources is \textit{Universalism: Concern}, with the least frequent being \textit{Hedonism} and \textit{Humility}.
In Figure \ref{level2-radars}(a), we can observe that the categories are similarly distributed across the main dataset splits, with some minor exceptions which can be attributed to the fact that IBM-ArgQ-Rank-30kArgs is the main source of arguments in our dataset and we ensured that no same conclusion occurs in different splits. When it comes to individual data sources from the \textit{supplementary} evaluation splits,  since all of the supplementary datasets are unique in terms of genre and moral reasoning, it is also reflected in the distribution of value categories within the arguments (cf. Figure \ref{level2-radars}b-d). Thus, \textit{Achievement} and \textit{Security: Societal} categories manifest themselves in the question-answering forum dataset, Zhihu. The NYT part also reflects value categories specific to the topics covered in it, with \textit{Security: Personal} appearing in more than 30\% of the arguments. In contrast, Nahj al-Balagha appears to be the most balanced data subset in terms of value categories.
Despite the described similarities and differences, we do not claim any of the parts as representative of the respective culture. In this case, we can only state that these distributions are descriptive of our dataset.  

\paragraph{Baseline experiments}
To assess the impact of dataset extension, we used the classification approaches listed in \citep{kiesel:2022b}. We trained and tested the models on the respective splits of the \textit{main} dataset. In comparison to the Webis-ArgValues-22, the effectiveness of a 1-Baseline (assigns each value to all of the arguments) decreases but that of an out-of-the-box BERT model increases across all evaluation metrics. A comparison of different evaluation metrics on the two datasets is demonstrated in Table \ref{table-compare-results}. Therefore, although the classification difficulty increased as per the label distribution, the larger dataset allows for training better models. 

\section{Conclusion}

We presented the Touch{\'e}23-ValueEval Dataset for Identifying Human Values behind Arguments, comprising 9\,324 arguments manually labelled for 54 values and 20 value categories. We detailed its construction and its complementary nature to the Webis-ArgValues-22 dataset. We expanded the previous dataset in terms of argument count, cultural variety, and writing style.
Finally, we reported baseline classification results that suggest that the expansion of the dataset allows for better learning of concepts by a vanilla BERT model. We hope that this dataset allows for more elaborate approaches for successful value detection, even beyond the ValueEval'23 task. 

\section{Ethics Statement}
\label{ethics}

Since this work is a direct continuation of our earlier work \cite{kiesel:2022b}, the same statement applies and we repeat it here for completeness.

Identifying values in argumentative texts could be used in various applications like argument faceted search, value-based argument generation, and value-based personality profiling. In all these applications, an analysis of values has the opportunity to broaden the discussion (e.g., by presenting a diverse set of arguments covering a wide spectrum of personal values in search or inviting people with underrepresented value-systems to discussions). At the same time, a value-based analysis could risk to exclude people or arguments based on their values. However, in other cases, for example hate speech, such an exclusion might be desirable.

While we tried to include texts from different cultures in our dataset, it is important to note that these samples are not representative of their respective culture, but intended as a benchmark for measuring classification robustness across sources. A more significant community effort is needed to collect more solid datasets from a wider variety of sources. To facilitate the inclusivity of different cultures, we adopted a personal value taxonomy that has been developed targeting universalism and tested across cultures. However, in our study, the annotations have all been carried out by annotators from a Western background. Even though the value taxonomy strives for universalism, a potential risk is that an annotator from a specific culture might fail to correctly interpret the implied values in a text written by people from a different culture.

Finally, we did not gather any personal information in our annotation studies, and we ensured that all our annotators get paid more than the minimum wage in the U.S.

 % \section*{Acknowledgements}

\bibliography{lit}

\begin{thebibliography}{6}
\expandafter\ifx\csname natexlab\endcsname\relax\def\natexlab#1{#1}\fi

\bibitem[{Barriere et~al.(2022)Barriere, Jacquet, and
  Hemamou}]{barriere-etal-2022-cofe}
Valentin Barriere, Guillaume~Guillaume Jacquet, and Leo Hemamou. 2022.
\newblock \href {https://aclanthology.org/2022.aacl-short.52} {{C}o{FE}: A new
  dataset of intra-multilingual multi-target stance classification from an
  online {E}uropean participatory democracy platform}.
\newblock In \emph{Proceedings of the 2nd Conference of the Asia-Pacific
  Chapter of the Association for Computational Linguistics and the 12th
  International Joint Conference on Natural Language Processing (Volume 2:
  Short Papers)}, pages 418--422, Online only. Association for Computational
  Linguistics.

\bibitem[{Farahani et~al.(2021)Farahani, Gharachorloo, Farahani, and
  Manthouri}]{farahani:2021}
Mehrdad Farahani, Mohammad Gharachorloo, Marzieh Farahani, and Mohammad
  Manthouri. 2021.
\newblock \href {https://doi.org/10.1007/s11063-021-10528-4} {{ParsBERT:
  Transformer-based model for Persian language understanding}}.
\newblock \emph{Neural Processing Letters}.

\bibitem[{Gretz et~al.(2020)Gretz, Friedman, Cohen{-}Karlik, Toledo, Lahav,
  Aharonov, and Slonim}]{gretz:2020}
Shai Gretz, Roni Friedman, Edo Cohen{-}Karlik, Assaf Toledo, Dan Lahav, Ranit
  Aharonov, and Noam Slonim. 2020.
\newblock \href {https://doi.org/10.1609/aaai.v34i05.6285} {A large-scale
  dataset for argument quality ranking: Construction and analysis}.
\newblock In \emph{34th {AAAI} Conference on Artificial Intelligence ({AAAI}
  2020)}, pages 7805--7813. {AAAI} Press.

\bibitem[{Hovy et~al.(2013)Hovy, Berg-Kirkpatrick, Vaswani, and
  Hovy}]{hovy:2013}
Dirk Hovy, Taylor Berg-Kirkpatrick, Ashish Vaswani, and Eduard Hovy. 2013.
\newblock Learning whom to trust with mace.
\newblock In \emph{Conference of the North American Chapter of the Association
  for Computational Linguistics: Human Language Technologies (NAACL-HLT 2013)},
  pages 1120--1130. Association for Computational Linguistics.

\bibitem[{Kiesel et~al.(2022)Kiesel, Alshomary, Handke, Cai, Wachsmuth, and
  Stein}]{kiesel:2022b}
Johannes Kiesel, Milad Alshomary, Nicolas Handke, Xiaoni Cai, Henning
  Wachsmuth, and Benno Stein. 2022.
\newblock \href {https://doi.org/10.18653/v1/2022.acl-long.306} {{Identifying
  the Human Values behind Arguments}}.
\newblock In \emph{60th Annual Meeting of the Association for Computational
  Linguistics (ACL 2022)}, pages 4459--4471. Association for Computational
  Linguistics.

\bibitem[{Schwartz(1994)}]{schwartz:1994}
Shalom~H Schwartz. 1994.
\newblock \href {https://doi.org/10.1111/j.1540-4560.1994.tb01196.x} {Are there
  universal aspects in the structure and contents of human values?}
\newblock \emph{Journal of Social Issues}, 50:19--45.

\end{thebibliography}

\appendix
\section{Annotation Interface}
\label{annotation-interface}

\begin{figure}
\centering\small
Complete\\
\includegraphics[width=.78\linewidth]{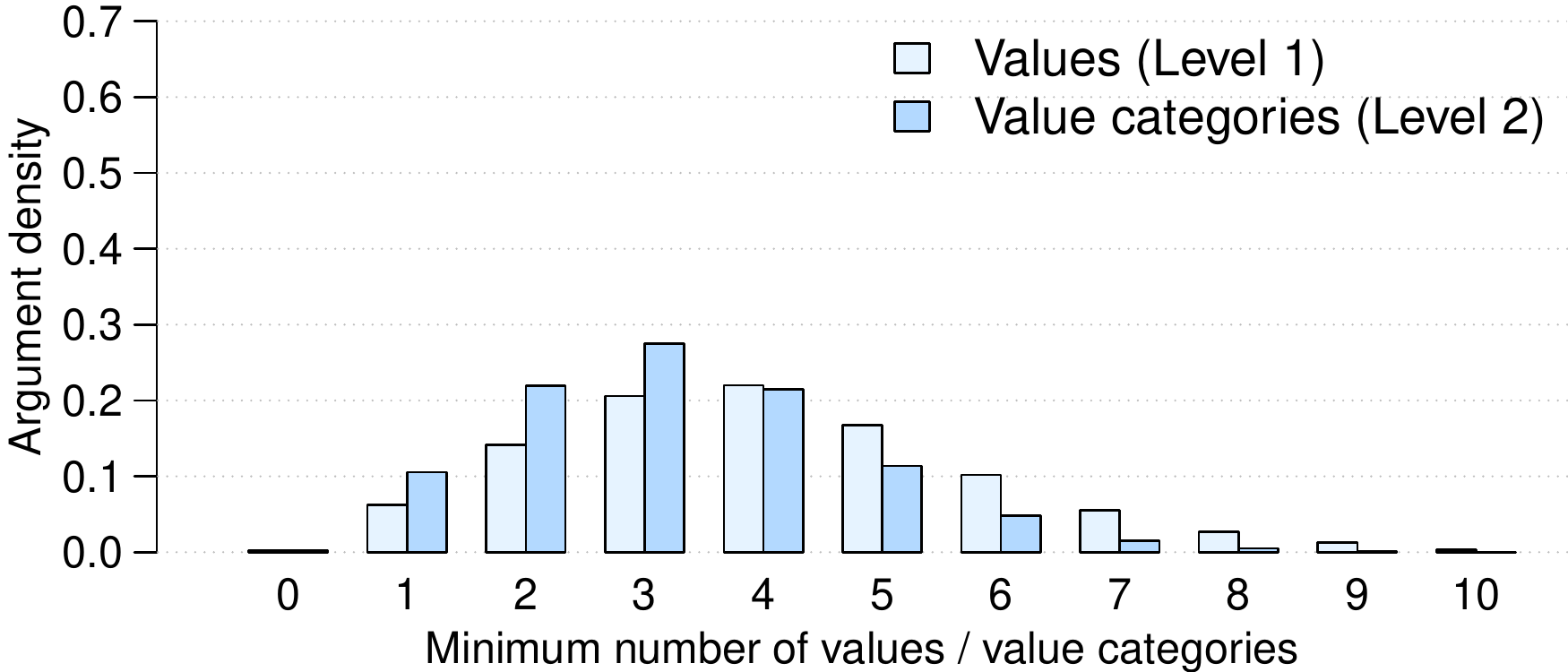}\\[2ex]
IBM-ArgQ-Rank-30kArgs\\
\includegraphics[width=.78\linewidth]{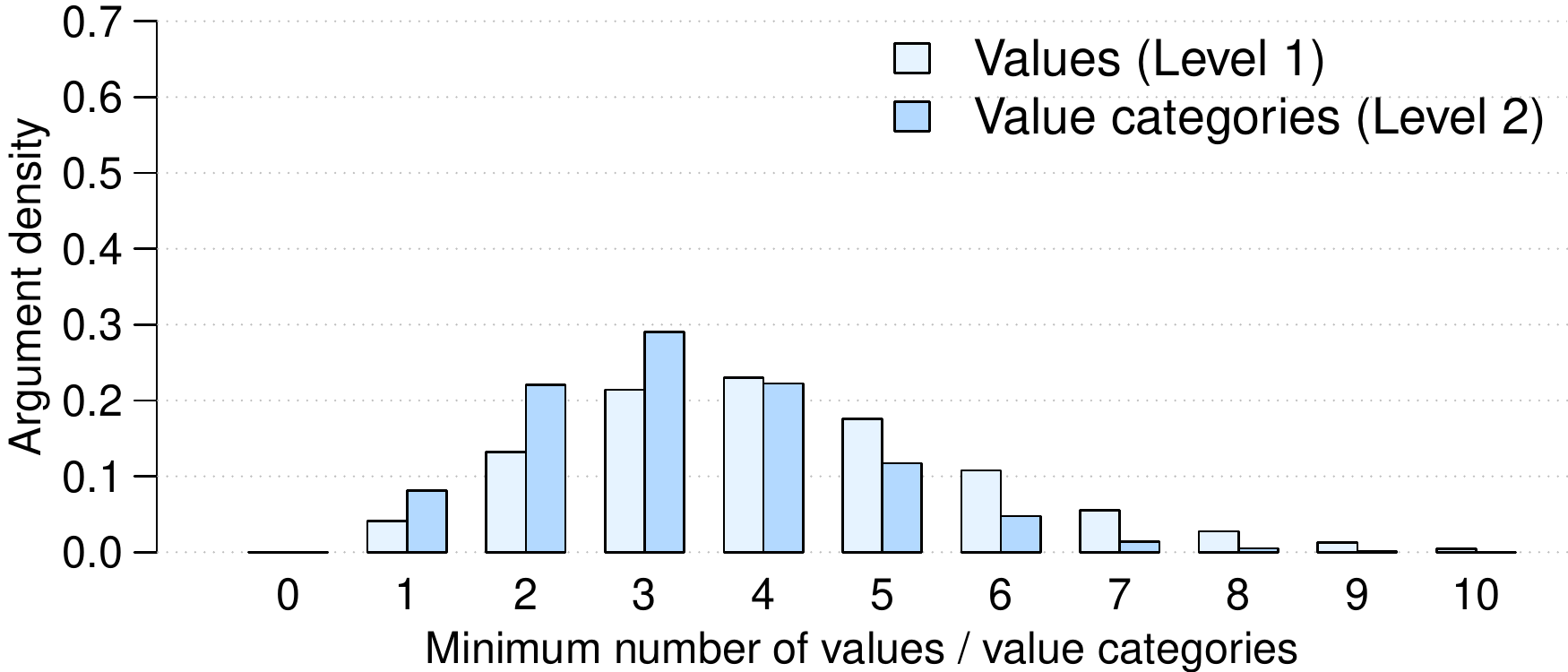}\\[2ex]
Conf.\ on the Future of Europe\\
\includegraphics[width=.78\linewidth]{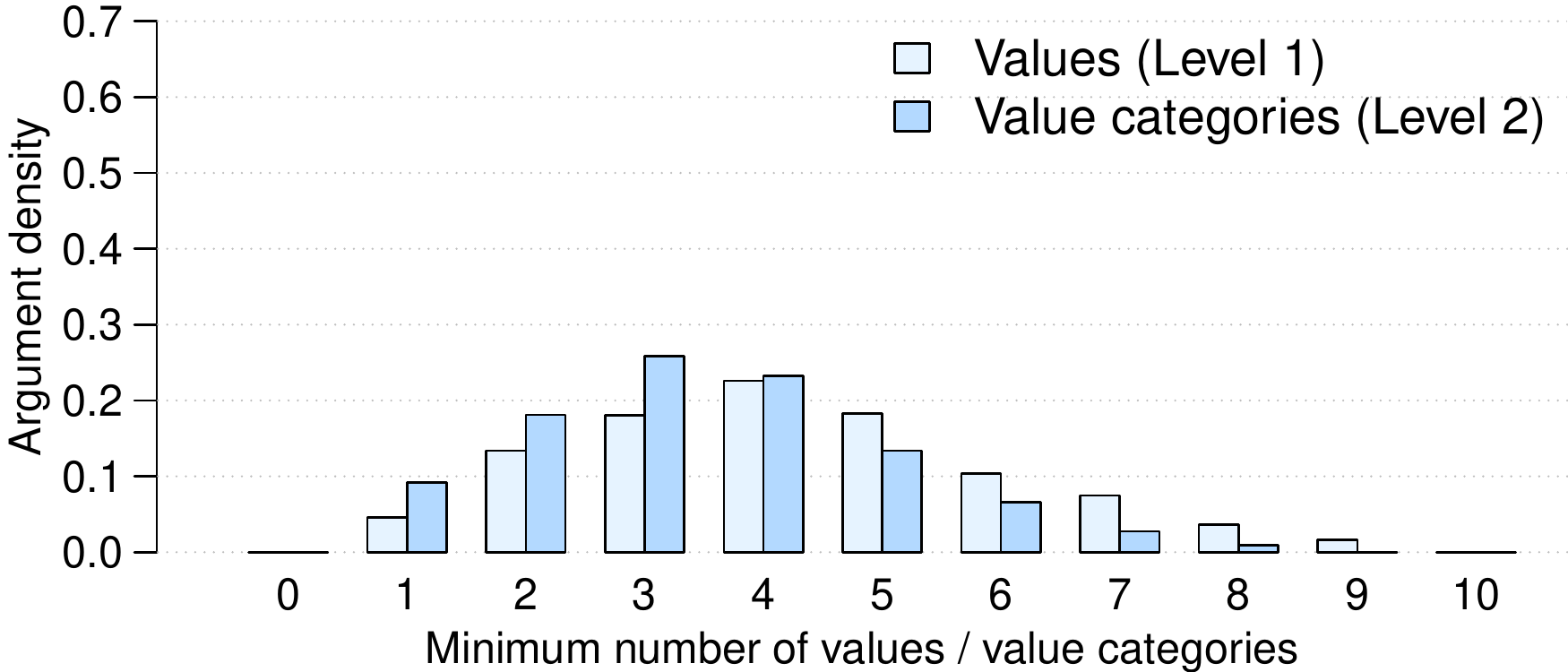}\\[2ex]
Group Discussion Ideas\\
\includegraphics[width=.78\linewidth]{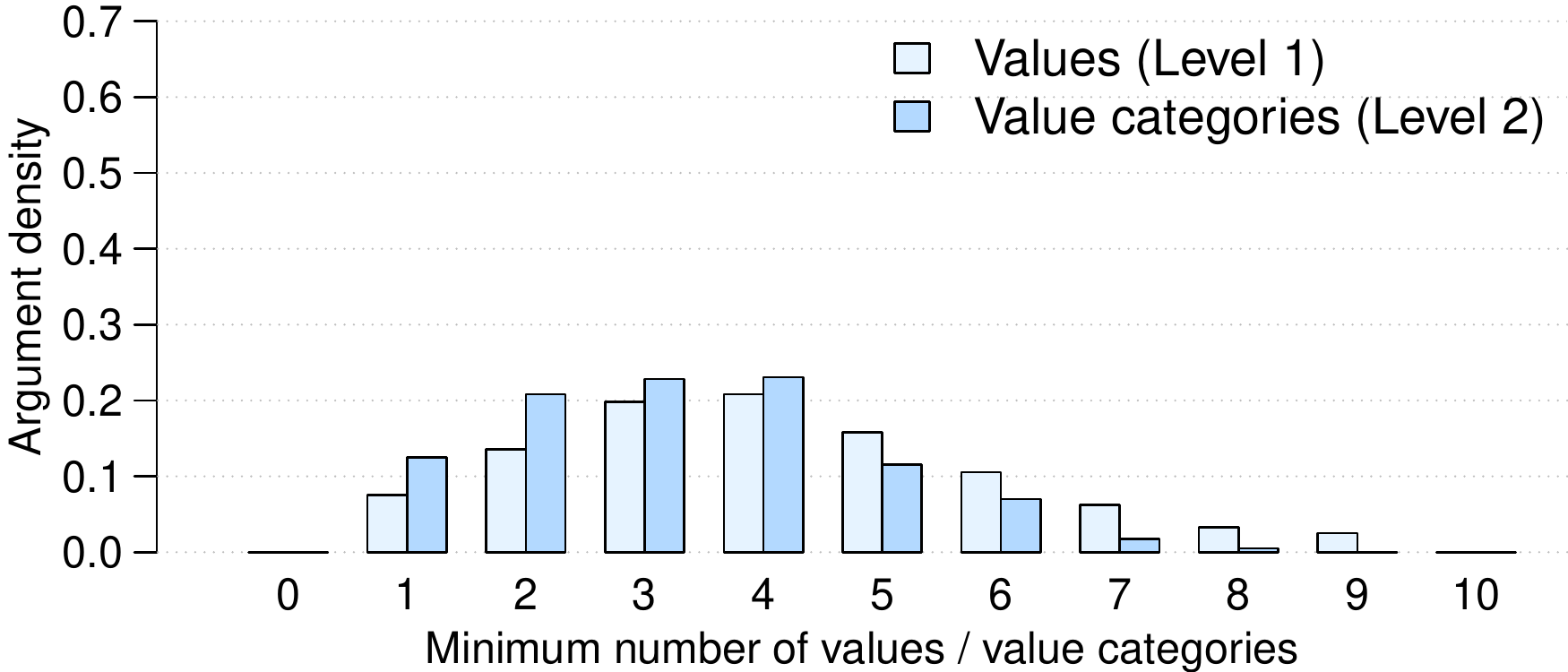}\\[2ex]
Zhihu\\
\includegraphics[width=.78\linewidth]{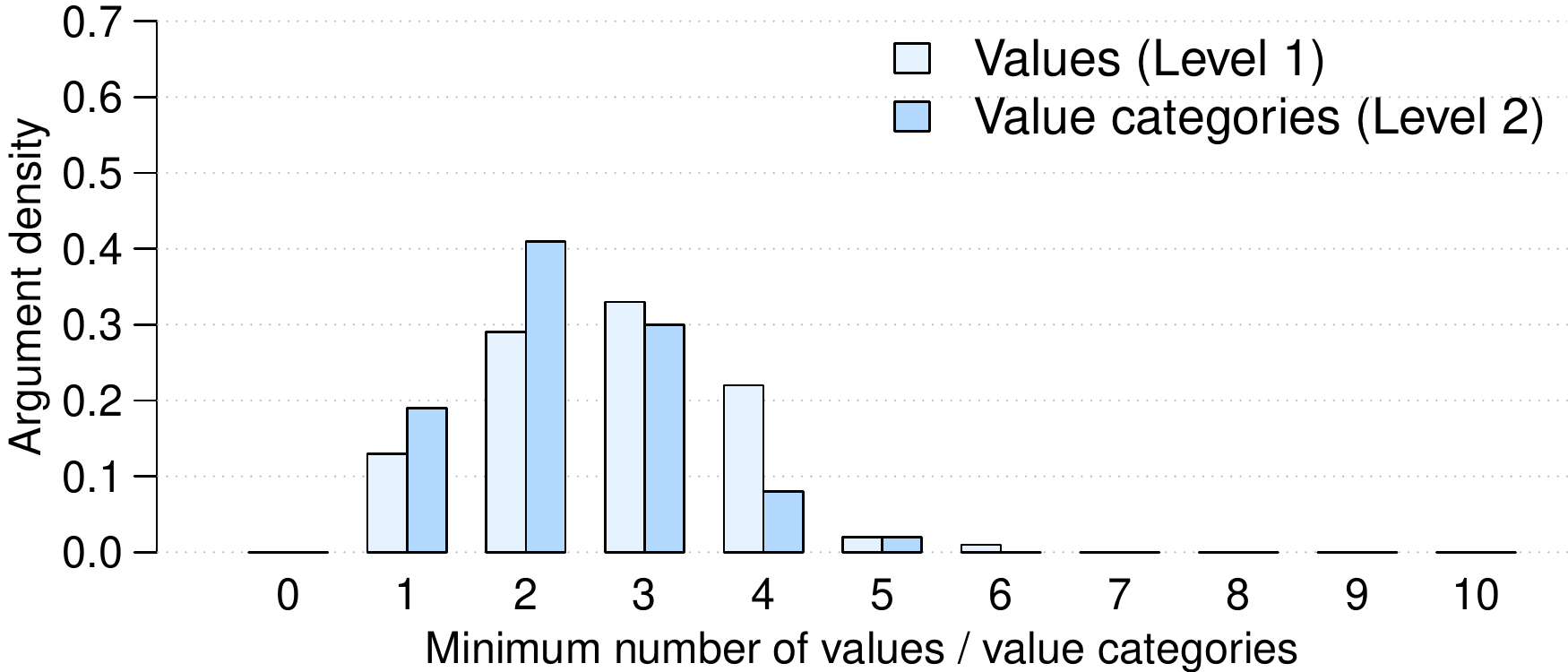}\\[2ex]
Nahj al-Balagha\\
\includegraphics[width=.78\linewidth]{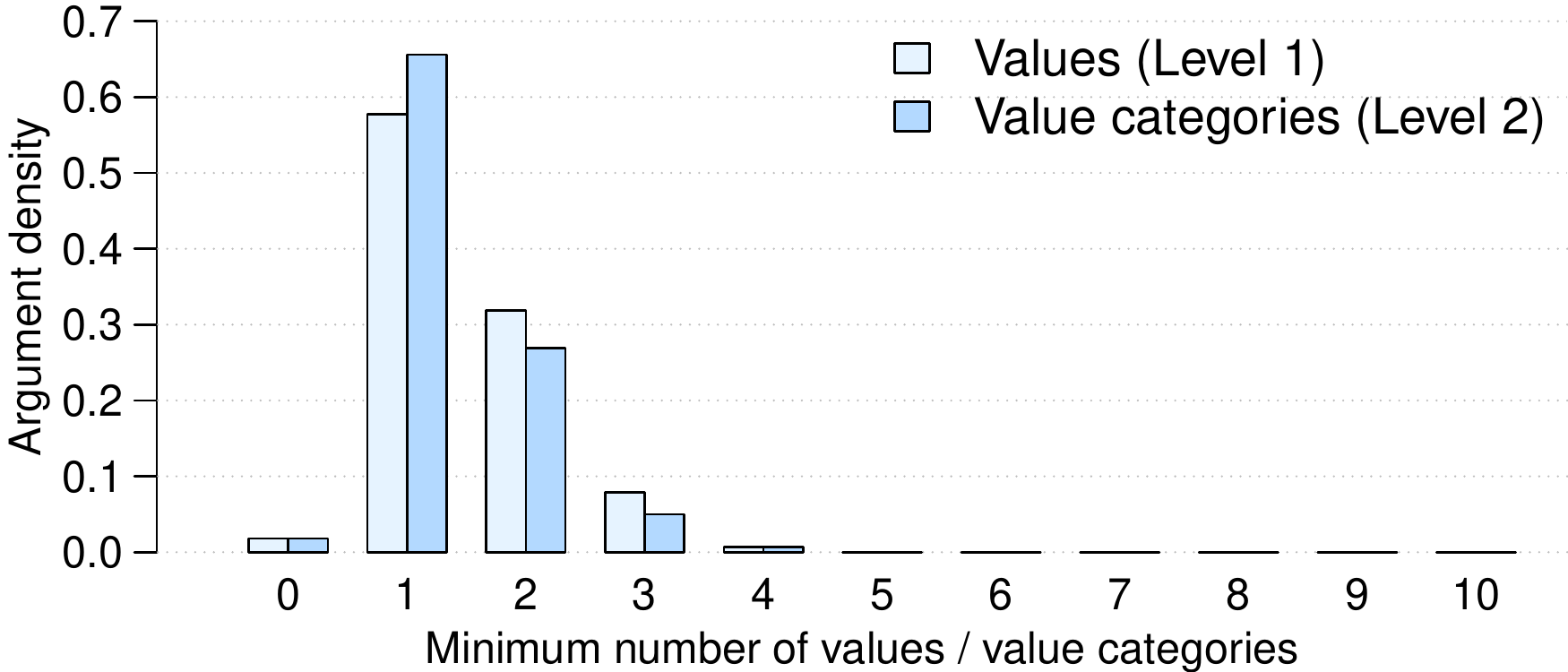}\\[2ex]
The New York Times\\
\includegraphics[width=.78\linewidth]{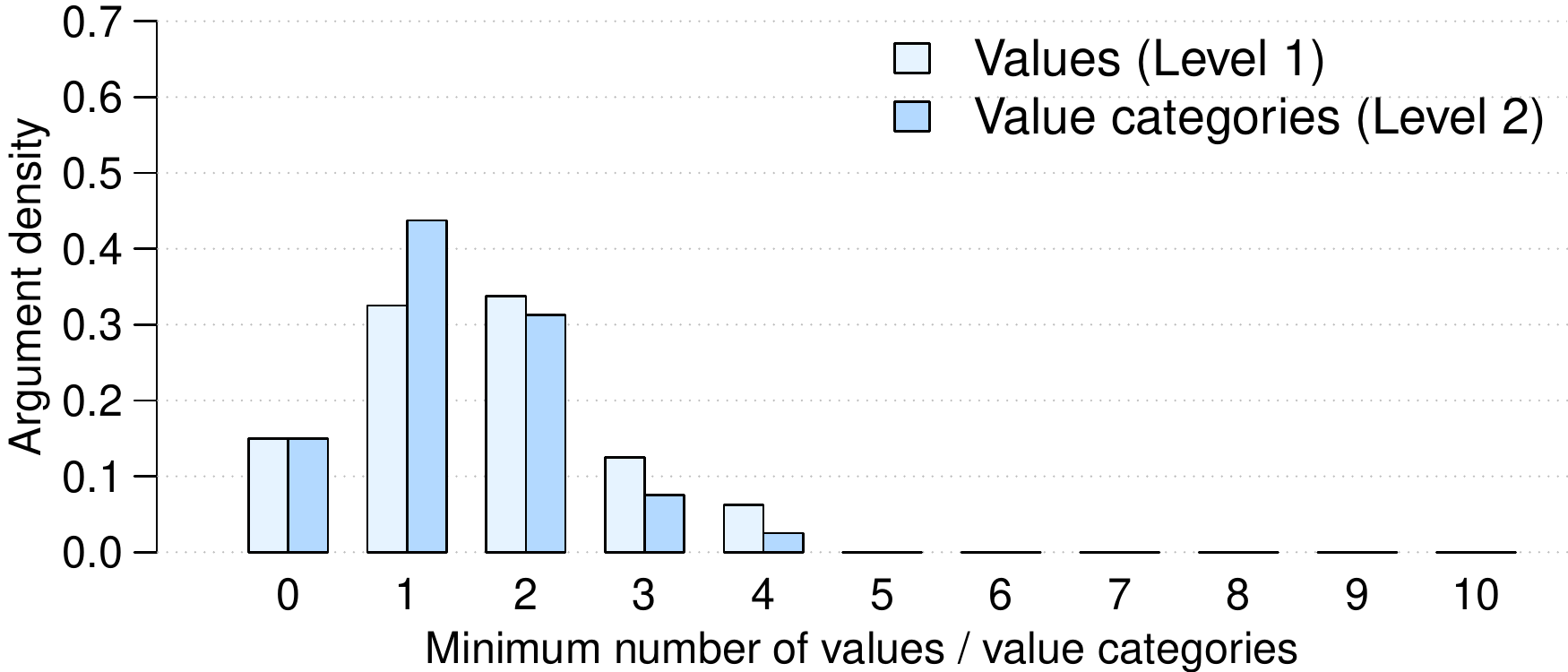}
\caption{Fraction of arguments per dataset part having a specific number of assigned values (out of~54) or value categories (out of~10).}
\label{density-number-of-labels}
\end{figure}

Figure~\ref{density-number-of-labels} shows the label distribution to allow for a comparison with Figure~2 from \citet{kiesel:2022b}.

\begin{figure*}
\includegraphics[width=\linewidth]{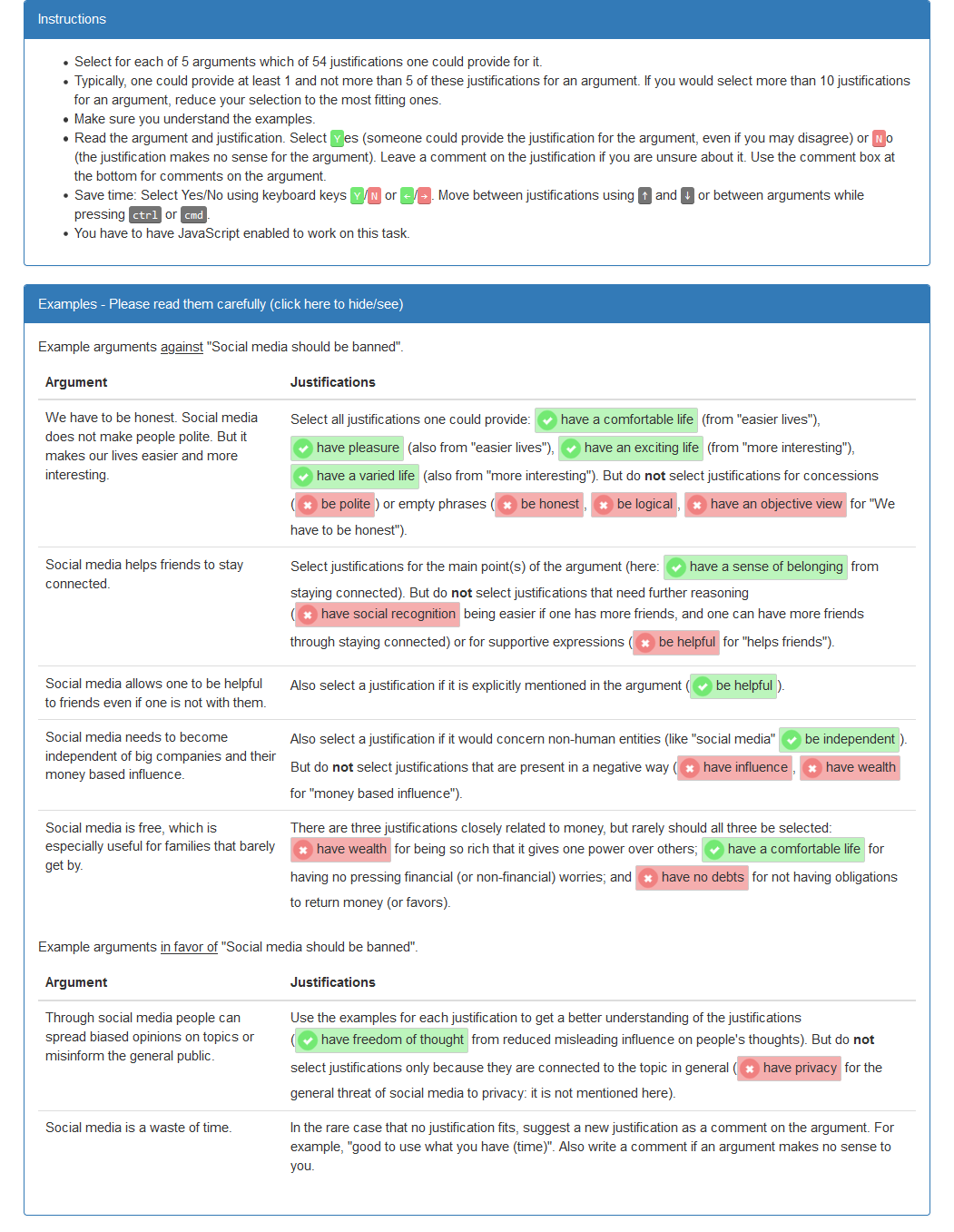}
\caption{Screenshot ot the first part of the annotation interface, containing instructions and examples.}
\label{argument-values-annotation-interface-top}
\end{figure*}

\begin{figure*}
\includegraphics[width=\linewidth]{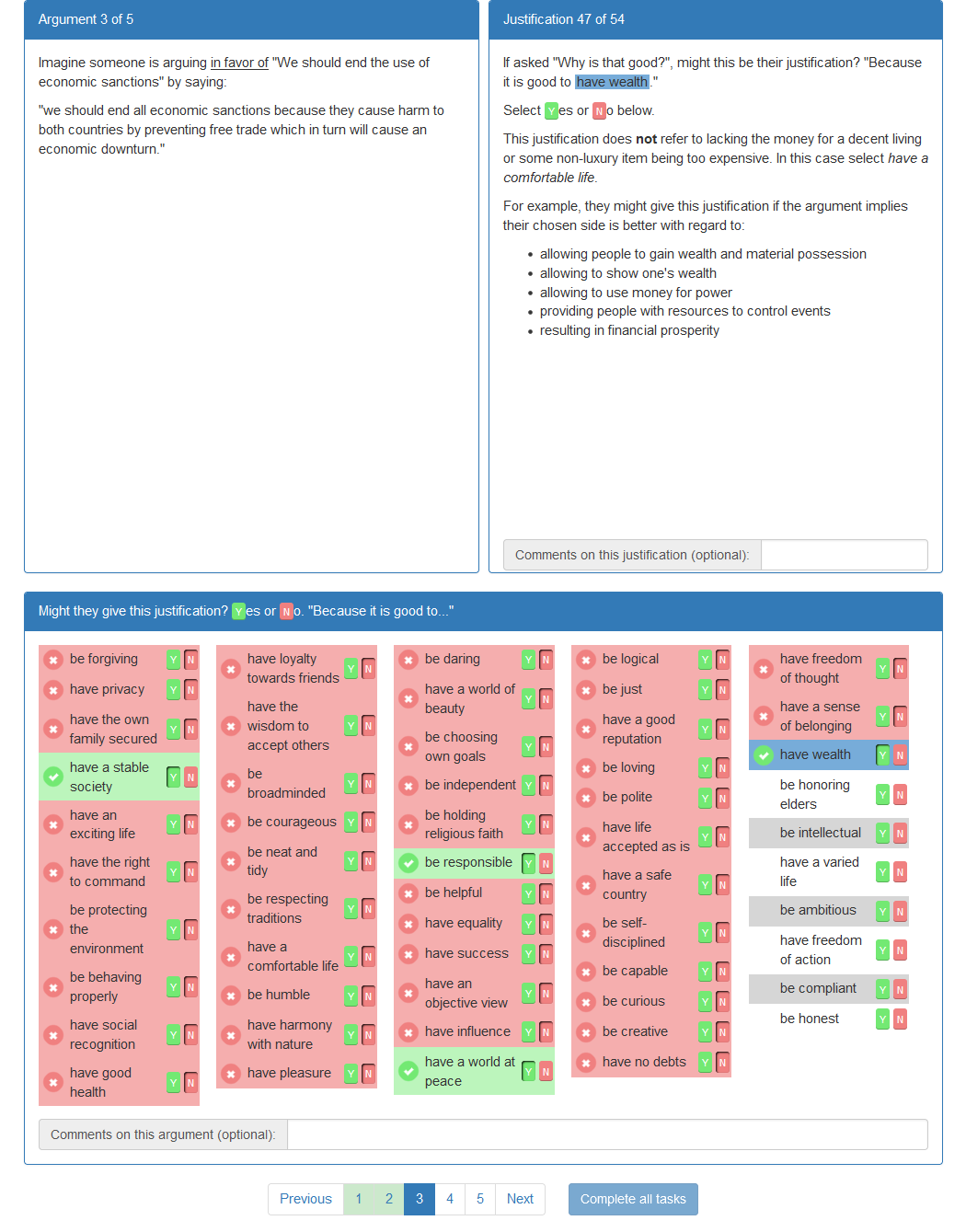}
\caption{Screenshot of the second part of the annotation interface, which consists of three panels:
(1)~the top left panel places the argument in a scenario (``Imagine'');
(2)~the top right panel formulates the annotation task for a value (here: {\em have wealth}) as a yes/no question, describing the value with examples; and
(3)~the bottom panel shows the annotation progress for the argument and allows for a quick review of selected annotations.}
\label{argument-values-annotation-interface-bottom}
\end{figure*}

Figures~\ref{argument-values-annotation-interface-top} and~\ref{argument-values-annotation-interface-bottom} show screenshots of the custom annotation interface taken from \citet{kiesel:2022b}. Its source code is distributed as part of the Webis-ArgValues-22 dataset at \url{https://github.com/webis-de/ACL-22}.

\end{document}